\documentclass[10pt]{article}
\usepackage[margin=1in]{geometry}
\usepackage[T1]{fontenc}
\usepackage{lmodern}
\usepackage{microtype}
\usepackage{amsmath,amssymb}
\usepackage{booktabs}
\usepackage{graphicx}
\usepackage[table]{xcolor}
\usepackage{tikz}
\usetikzlibrary{arrows.meta,positioning,fit,backgrounds}
\usepackage[numbers,sort&compress]{natbib}

\usepackage{enumitem}
\usepackage[hidelinks]{hyperref}

\definecolor{sig}{HTML}{DCEBF7}

\newcommand{\code}[1]{\texttt{#1}}
\newcommand{\SourceCommit}{de48cdd}
\newcommand{\SourceRef}{v0.1.0}
\newcommand{\NDags}{20}
\newcommand{\NReal}{3}
\newcommand{\NRuns}{60}

\newcommand{\MkGainPct}{45}
\newcommand{\CostGainPct}{11}
\newcommand{\MkDiffHeft}{98.8}
\newcommand{\MkDiffHeftLo}{59.9}
\newcommand{\MkDiffHeftHi}{140.6}

\newcommand{\BvDiffHeft}{0.35}
\newcommand{\BvDiffHeftLo}{0.20}
\newcommand{\BvDiffHeftHi}{0.50}
\newcommand{\RlHeftFewer}{0.53}
\newcommand{\RlHeftFewerLo}{0.10}
\newcommand{\RlHeftFewerHi}{1.05}
\newcommand{\MkWinsHeft}{17}
\newcommand{\MkLossHeft}{2}
\newcommand{\MkTieHeft}{1}
\newcommand{\CostWinsHeft}{19}
\newcommand{\TiePct}{0.5}
\newcommand{\MkDiffCheap}{44.6}

\newcommand{\RlDiffCheap}{9.2}

\newcommand{\SuccDiffRouter}{0.026}
\newcommand{\SuccDiffRouterLo}{0.005}
\newcommand{\SuccDiffRouterHi}{0.054}
\newcommand{\SuccOurs}{0.974}
\newcommand{\SuccRouter}{1.000}
\newcommand{\RouterCostX}{9}
\newcommand{\RouterMkX}{1.5}
\newcommand{\RouterRl}{38}
\newcommand{\RlOurs}{1.32}
\newcommand{\BvOurs}{0.38}
\newcommand{\StaticFaster}{24.6}
\newcommand{\StaticFasterLo}{8.8}
\newcommand{\StaticFasterHi}{42.6}

\newcommand{\StaticWins}{14}
\newcommand{\StaticUnplanned}{2.6}
\newcommand{\RiskZeroFaster}{9.6}
\newcommand{\RiskZeroFasterLo}{1.3}
\newcommand{\RiskZeroFasterHi}{19.2}

\newcommand{\PanicRunsWith}{19}
\newcommand{\PanicRunsWithout}{41}
\newcommand{\PanicGapWith}{61.4}
\newcommand{\PanicGapWithout}{7.6}

\newcommand{\SolvesOurs}{21}
\newcommand{\ResolvesOurs}{20}
\newcommand{\NSolves}{1{,}245}
\newcommand{\SolveMsPFifty}{20}
\newcommand{\SolveMsPNinetyFive}{163}
\newcommand{\SolveMsPNinetyNine}{770}
\newcommand{\SolveMsMax}{2{,}385}
\newcommand{\SolveSRunMean}{1.10}
\newcommand{\SolveSRunMax}{6.7}
\newcommand{\ChargedHeft}{97.7}
\newcommand{\ChargedHeftLo}{58.9}
\newcommand{\ChargedHeftHi}{139.3}
\newcommand{\ChargedStatic}{25.5}
\newcommand{\ChargedStaticLo}{9.7}
\newcommand{\ChargedStaticHi}{43.4}
\newcommand{\TpmAnthropic}{40{,}000}
\newcommand{\RpmAnthropic}{30}
\newcommand{\TpmOpenai}{60{,}000}
\newcommand{\RpmOpenai}{40}
\newcommand{\CalDate}{2026-11-29}
\newcommand{\CalProbes}{64, 256, 768}
\newcommand{\CalSamples}{6}
\newcommand{\CalRmseMax}{0.33}
\newcommand{\LiveTasks}{11}
\newcommand{\LiveDone}{11}
\newcommand{\LiveMakespan}{57.9}
\newcommand{\LiveCost}{0.0301}

\newcommand{\LiveTruncated}{0}
\newcommand{\LiveReasoning}{0}
\newcommand{\LiveRateLimited}{0}
\newcommand{\NOptimal}{1{,}064}
\newcommand{\NFeasible}{3}
\newcommand{\NRelaxed}{178}
\newcommand{\OptimalPct}{85}
\newcommand{\LlamaBefore}{3}
\newcommand{\LlamaAfter}{28}
\newcommand{\LlamaStatic}{8}
\newcommand{\NRelaxedDags}{12}
\newcommand{\TopFourShare}{65}
\newcommand{\DagR}{$-$0.45}
\newcommand{\DagRLo}{$-$0.85}
\newcommand{\DagRHi}{0.03}
\newcommand{\RestrictedHeft}{90.1}
\newcommand{\RestrictedHeftN}{49}
\newcommand{\RestrictedStatic}{30.1}
\newcommand{\RestrictedStaticN}{46}
\newcommand{\NIntervals}{47}
\newcommand{\LiveReSolves}{8}
\newcommand{\LiveRatioMin}{1.3}
\newcommand{\LiveRatioMax}{2.2}
\newcommand{\LiveModelsUsed}{2}
\newcommand{\FuRef}{v0.2.0}
\newcommand{\FuCommit}{aeb2c43}
\newcommand{\PreregCommit}{c57ed9b}
\newcommand{\FuDags}{80}
\newcommand{\FuReal}{3}
\newcommand{\FuRuns}{240}
\newcommand{\FuIntervals}{23}
\newcommand{\FuMkRoll}{130.8}
\newcommand{\FuCostRoll}{0.0967}

\newcommand{\SweepQualified}{7}
\newcommand{\SweepTied}{0, 0.25, 0.5, 1 and 2}
\newcommand{\SweepLambda}{2}
\newcommand{\FuMkHeft}{142.4}
\newcommand{\FuMkHeftLo}{117.6}
\newcommand{\FuMkHeftHi}{168.0}
\newcommand{\FuCostHeft}{0.0132}
\newcommand{\FuCostHeftLo}{0.0111}
\newcommand{\FuCostHeftHi}{0.0155}
\newcommand{\FuWinsHeft}{78}
\newcommand{\FuMkCheap}{42.6}

\newcommand{\FuMkRouter}{74.5}

\newcommand{\FuStaticFaster}{21.2}
\newcommand{\FuStaticFasterLo}{12.5}
\newcommand{\FuStaticFasterHi}{31.0}
\newcommand{\FuStaticOver}{0.0010}
\newcommand{\FuStaticOverLo}{0.0003}
\newcommand{\FuStaticOverHi}{0.0017}

\newcommand{\FuPricedFaster}{15.2}
\newcommand{\FuPricedFasterLo}{8.5}
\newcommand{\FuPricedFasterHi}{23.2}
\newcommand{\FuPricedOver}{0.00042}
\newcommand{\FuPricedOverLo}{0.00028}
\newcommand{\FuPricedOverHi}{0.00059}

\newcommand{\FuPricedSuccPts}{0.35}
\newcommand{\FuPricedSuccPtsLo}{0.02}
\newcommand{\FuPricedSuccPtsHi}{0.89}
\newcommand{\FuPricedBetter}{36}
\newcommand{\FuPricedWorse}{4}
\newcommand{\FuPricedTie}{40}
\newcommand{\FuPricedShare}{72}
\newcommand{\FuPricedShareLo}{52}
\newcommand{\FuPricedShareHi}{92}
\newcommand{\FuPaidMk}{$-$1.4}
\newcommand{\FuPaidMkLo}{$-$4.0}
\newcommand{\FuPaidMkHi}{1.6}

\newcommand{\OldPaidMk}{$-$10.3}
\newcommand{\OldPaidMkLo}{$-$24.0}
\newcommand{\OldPaidMkHi}{$-$2.0}
\newcommand{\OldPricedFaster}{16.6}
\newcommand{\FuRiskZeroSuccPts}{$-$1.0}
\newcommand{\FuRiskZeroSuccPtsLo}{$-$1.9}
\newcommand{\FuRiskZeroSuccPtsHi}{$-$0.3}

\newcommand{\PhStaticFaster}{5.9}
\newcommand{\PhStaticFasterLo}{1.4}
\newcommand{\PhStaticFasterHi}{11.6}

\newcommand{\PhSuccPts}{1.3}
\newcommand{\PhSuccPtsLo}{0.5}
\newcommand{\PhSuccPtsHi}{2.4}
\newcommand{\FuPanicWith}{95}
\newcommand{\FuPanicWithout}{145}
\newcommand{\FuPanicGapWith}{47.9}
\newcommand{\FuPanicGapWithout}{3.6}
\newcommand{\FuDagR}{$-$0.38}
\newcommand{\FuDagRLo}{$-$0.55}
\newcommand{\FuDagRHi}{$-$0.26}
\newcommand{\FuRerunCommit}{75022a9}
\newcommand{\FuSolveMean}{1.3}
\newcommand{\FuSolveMax}{27.6}
\newcommand{\FuSolveOverTen}{6}
\newcommand{\FuSolveChargedHeft}{141.1}
\newcommand{\FuSolveChargedHeftLo}{116.3}
\newcommand{\FuSolveChargedHeftHi}{166.7}
\newcommand{\FuSolveChargedStatic}{22.2}
\newcommand{\FuSolveChargedStaticLo}{13.4}
\newcommand{\FuSolveChargedStaticHi}{32.2}
\newcommand{\FuSolveRerunMean}{1.9}
\newcommand{\FuSolveRerunMax}{31.9}
\newcommand{\FuSolveRerunOverTen}{9}
\newcommand{\FuSolveRerunChargedHeft}{140.5}
\newcommand{\FuSolveRerunChargedHeftLo}{115.8}
\newcommand{\FuSolveRerunChargedHeftHi}{166.0}
\newcommand{\FuSolveRerunChargedStatic}{22.6}
\newcommand{\FuSolveRerunChargedStaticLo}{13.9}
\newcommand{\FuSolveRerunChargedStaticHi}{32.6}
\newcommand{\FuSolveWorstMk}{98}

\title{Planning Ahead Under Rate Limits:\\
Rolling-Horizon CP-SAT Scheduling for Heterogeneous LLM Agent Fleets}
\author{Pouria Mortezaagha\qquad Arya Rahgozar\\
\small University of Ottawa\\
\small \texttt{\{pmort101,\,arahgoza\}@uottawa.ca}\\
\small \url{https://github.com/pouriamrt/neuro-orch}}
\date{}

\begin{document}
\maketitle

\begin{abstract}
Multi-step LLM workflows increasingly mix models from several providers.
Each model has its own skills, price, speed, context window and reliability, and providers cap the tokens and requests a client may send per minute.
Deciding which model runs each task and when is therefore a scheduling problem, yet most orchestrators decide one call at a time.
We formulate it as a constraint program solved with OR-Tools CP-SAT.
Each feasible (task, model) pair is an optional interval, and a second, 60-second interval per pair turns these per-minute limits into two cumulative constraints without time buckets.
Precedence, concurrency and the budget are constraints (the budget is relaxed only when it cannot be met), and the objective trades makespan, cost and risk.
Because LLM calls drift from their estimates, fail and spawn follow-up work, a rolling-horizon engine re-solves the remaining work when execution departs from the plan.
On \NDags{} held-out random workflows with \NReal{} stochastic realizations each, in a seeded simulator with common random numbers, the scheduler finishes \MkDiffHeft\,s sooner than a capacity-aware online HEFT baseline (95\% CI [\MkDiffHeftLo, \MkDiffHeftHi]; a \MkGainPct\% reduction) and costs \CostGainPct\% less, with a median solve time of \SolveMsPFifty\,ms (mean total \SolveSRunMean\,s per run).
Re-planning, however, is not free: solving once and never re-planning is faster still, by \StaticFaster\,s (CI [\StaticFasterLo, \StaticFasterHi]), with no detectable difference in cost.
A post-hoc analysis places most of that gap in runs whose re-solves had to relax the budget, which shifts work onto a free but single-slot local model.
On \FuDags{} fresh workflows, a pre-registered change that prices the overspend instead of minimizing it first recovers about \FuPricedShare\% of the single solve's lead on those workflows at a small extra spend, which supports this account, and the advantage over HEFT holds.
A reliability-first router completes more tasks at about \RouterCostX$\times$ the cost.
Code and data are released.
\end{abstract}

\section{Introduction}

A research-report workflow might plan an outline, run several searches, verify facts, write sections and summarize.
Each step can go to a different model: a frontier model for reasoning, a small model for summaries, a local model for cheap checks.
Choosing the model per task is only half the problem.
The calls also compete for shared capacity.
Providers limit the tokens and requests a client may send per minute, every model allows only so many requests in flight, the workflow has a budget, and tasks wait on their parents.
A choice that is cheapest for one task can stall three others behind a saturated rate window.

Current practice handles these pieces separately.
Model routers and cascades pick a model per query to trade quality against price~\citep{chen2023frugalgpt,ong2024routellm,ding2024hybridllm}.
Agent frameworks run a task graph and dispatch whatever is ready~\citep{wu2023autogen,hong2023metagpt,kim2023llmcompiler}.
Client libraries typically meet rate limits by backing off and retrying after HTTP 429 responses, and recent systems add admission control in front of rate-limited APIs~\citep{kosorin2026dispatch,agyemang2026hivemind} or allocate models under budgets and deadlines~\citep{wang2026ontime,yang2025bamas}.
We did not find a system that chooses models, timing and budget jointly in one constraint model that also encodes the rate limits shared across tasks (Section~\ref{sec:related}).

Classical scheduling has the right tools for this.
Assigning tasks with precedence to heterogeneous machines with capacity limits is a multi-mode resource-constrained project scheduling problem~\citep{hartmann2010rcpsp}, and constraint programming solvers such as CP-SAT~\citep{ortools,cpsat,perron2023cpsatlp} handle such models well.
LLM workloads add two complications.
First, provider limits cap tokens and requests per minute over continuous time, not per fixed period.
Second, durations and token counts are only estimates: the realized values drift, calls fail, and some tasks create new tasks.

This paper asks whether planning with a constraint model pays off against a strong heuristic once these complications are present, and whether re-planning during execution helps.
We contribute:
\begin{itemize}[leftmargin=*,itemsep=2pt]
  \item A CP-SAT model for assigning and timing LLM tasks on a heterogeneous fleet (Section~\ref{sec:model}). It encodes tokens-per-minute (TPM) and requests-per-minute (RPM) limits as cumulative constraints over 60-second request intervals, exactly under request-time counting, and reports which soft constraint (budget or deadline) blocks feasibility through a lexicographic relaxation.
  \item A rolling-horizon engine that re-solves on duration deviation, failure and task spawning, with deterministic single-worker solves so that runs are reproducible (Section~\ref{sec:rolling}).
  \item An evaluation protocol built to be attacked (Section~\ref{sec:setup}): common random numbers across strategies, disjoint tuning and reporting seeds, a capacity-aware online version of HEFT (Heterogeneous Earliest Finish Time)~\citep{topcuoglu2002heft} rather than a strawman, a DAG-level paired bootstrap, and a results file that is byte-identical on rerun.
  \item Results that include where the method loses (Section~\ref{sec:results}). Planning beats online HEFT on makespan and cost, but in our setting re-planning makes the scheduler slower than a single up-front solve, and a post-hoc analysis links much of that gap to how re-solves react to budget overruns.
A follow-up on fresh workflows with two pre-registered changes to the budget handling supports that link, and the advantage over HEFT holds there too (Section~\ref{sec:followup}).
\end{itemize}

\section{Problem setting}
\label{sec:problem}

A workflow is a DAG $G=(T,E)$.
Task $t\in T$ needs a skill set $S_t$, has estimated input and output tokens $i_t,o_t$ (total $k_t=i_t+o_t$), a priority $\pi_t\in\{1,\dots,5\}$ and a minimum acceptable reliability $\rho_t$.
An edge $(u,t)\in E$ means $t$ may start only after $u$ ends.
The workflow may carry a budget $B$ and a deadline $D$.

The fleet is a set of agents $A$, each one model.
Agents that share a rate limit form a pool $p(a)\in P$, which we call a provider.
Agent $a$ has skills, a per-skill reliability $r_{a,s}$, a context window $C_a$, a concurrency cap $\kappa_a$, a fixed latency $\lambda_a$, a decoding speed $\tau_a$ in tokens per second, and per-token prices.
Its reliability on task $t$ is that of its weakest required skill, $r_{at}=\min_{s\in S_t} r_{a,s}$.
Its estimated duration and cost on $t$ are
\begin{equation}
  d_{at}=\max\!\big(1,\left\lceil \lambda_a + o_t/\tau_a \right\rceil\big), \qquad
  c_{at}=i_t\,\mathrm{price}^{\mathrm{in}}_a + o_t\,\mathrm{price}^{\mathrm{out}}_a .
\end{equation}
with $c_{at}$ rounded to integer micro-dollars.
Provider $p$ allows at most $\mathrm{TPM}_p$ tokens and $\mathrm{RPM}_p$ requests in any 60-second window, shared by all its agents.
Real providers set limits in different ways: Anthropic gives each model class its own limits~\citep{anthropicratelimits}, while OpenAI's limits vary by model and some model families share one~\citep{openairatelimits}.
A per-model limit is a provider with one agent, and a shared family is a provider with several.
CP-SAT works on integers, so time is in seconds, money in micro-dollars, and probabilities are scaled by $10^4$.

At run time, realized tokens and durations differ from the estimates, an attempt fails with probability $1-r_{at}$, a request that exceeds the context window fails without being billed, a request the provider cannot admit returns HTTP 429, and some tasks spawn children.
These are the outcomes our simulator generates (Section~\ref{sec:setup}); against real APIs the same outcomes come from the client library's error responses.
A scheduler must produce, and keep revising, an assignment of each task to an agent and a start time.
The main measure is the makespan, the time until the workflow's last task ends.

\section{The CP-SAT model}
\label{sec:model}

\paragraph{Eligibility.}
Before building the model, a filter keeps pair $(t,a)$ only if $S_t\subseteq\mathrm{skills}(a)$, $C_a\ge 1.2\,k_t$, $r_{at}\ge\rho_t$, and $k_t\le \mathrm{TPM}_{p(a)}$.
A task with no eligible agent is reported with the reason for each agent and never reaches the solver.
Let $\mathcal{E}$ be the eligible pairs.

\paragraph{Assignment and time.}
Each task has start and end variables $s_t,e_t$.
Each eligible pair has a Boolean $x_{at}$ and an optional interval $I_{at}$ of length $d_{at}$ that is present iff $x_{at}=1$ and shares $s_t$ and $e_t$:
\begin{equation}
  \textstyle\sum_{a:(a,t)\in\mathcal{E}} x_{at}=1, \qquad
  x_{at}=1 \Rightarrow e_t=s_t+d_{at}.
\end{equation}
One model therefore chooses the agent and the start together, which keeps the trade-off between cost and makespan inside a single solve.
A two-stage approach (assign, then sequence) would lose it.

\paragraph{Precedence and concurrency.}
For every edge, $s_t\ge e_u$.
A finished parent enters with its actual end; a running parent enters with its estimated end rounded up, at least one second after its start, or one second from now if it is already overdue.
For each agent, a cumulative constraint with demand 1 and capacity $\kappa_a$ covers its optional intervals plus fixed intervals for its running tasks.

\paragraph{Rate limits.}
We assume request-time counting: a provider counts a request's tokens when the request is sent.
Under that rule, the load on provider $p$ at time $\theta$ is the sum of $k_t$ over requests sent in $(\theta-60,\theta]$.
We give each pair a second optional interval $W_{at}=[s_t,s_t+60)$ with the same presence literal $x_{at}$, and add per provider
\begin{equation}
  \textsc{Cumulative}\big(\{W_{at}\}_{p(a)=p}\cup R_p,\ \{k_t\},\ \mathrm{TPM}_p\big),\qquad
  \textsc{Cumulative}\big(\{W_{at}\}_{p(a)=p}\cup R_p,\ 1,\ \mathrm{RPM}_p\big),
\end{equation}
where $R_p$ holds fixed windows for requests sent in the last 60 seconds before the current solve, with their estimated tokens and their start times rounded up, so the limit carries over between re-solves.
A cumulative constraint~\citep{aggoun1993cumulative} bounds the summed demand of the intervals covering every time point, which here is exactly the sliding-window load.
The encoding needs no time buckets.
For the requests being planned it is exact under request-time counting on integer seconds, with token demands taken from the estimates $k_t$; rounding the carried windows' starts up makes them conservative.

\paragraph{Relation to token buckets.}
Real providers do not all meter this way. Anthropic documents a token bucket that is ``continuously replenished up to your maximum limit''~\citep{anthropicratelimits}. A sliding-window schedule is still safe under such a bucket:

\noindent\textbf{Proposition.} \emph{Consider a token bucket of capacity $L$, refilled at $L/60$ per second, full at time 0, that deducts a request's tokens when it is sent and admits a request only if the bucket holds at least that many. If the requests' tokens sum to at most $L$ in every half-open window $[\theta,\theta+60)$, the bucket admits every request.} The proof is in Appendix~\ref{app:proof}.

The converse fails: a bucket admits up to $2L$ tokens within 60 seconds when it starts full, so the window encoding can be up to twice as strict as the bucket and gives up some throughput. The guarantee also does not cover what the documentation adds on top of the bucket: separate limits for input and output tokens, with output counted as it is generated; enforcement over sub-minute intervals; and ramp limits on sudden bursts~\citep{anthropicratelimits}. OpenAI counts a request against TPM using the larger of \code{max\_tokens} and its token estimate~\citep{openairatelimits}. The guarantee also concerns the tokens the scheduler counts, which are estimates. Where limits are per model, a window shared by several models, as in our benchmark fleet, is more conservative still. Our simulator uses the sliding-window rule, so these differences are not exercised in the benchmark (Section~\ref{sec:limits}).

\paragraph{Budget and deadline.}
$\sum_{(a,t)\in\mathcal{E}} c_{at}x_{at}\le B'$, where $B'$ is the budget left after subtracting what has been spent and the estimated cost of running tasks.
The sum runs over pending tasks only.
A deadline, if given, bounds the makespan $M$, the latest end over pending and running tasks (and at least the current time).

\paragraph{Objective.}
With risk $R=\sum (1-r_{at})\,\pi_t\,x_{at}$ and cost $C=\sum c_{at}x_{at}$, both over pending tasks, the solver minimizes
\begin{equation}
  w_M\,\frac{M-\mathrm{now}}{L_M} + w_C\,\frac{C}{L_C} + w_R\,\frac{R}{L_R},
\end{equation}
where each $L$ is a lower bound computed before solving: the critical path using each task's fastest eligible duration, the sum of each task's cheapest eligible cost, and the sum of each task's lowest eligible risk, each floored at 1.
In the integer model each term's coefficient is $\mathrm{round}(10^9 w/L)$.
Dividing by lower bounds instead of loose upper bounds keeps the weights interpretable.
It also means a term with a tiny bound weighs more than its nominal weight, which mattered in tuning (Section~\ref{sec:setup}).
The default weights are $(w_M,w_C,w_R)=(0.45,0.45,0.1)$.

\paragraph{Infeasibility.}
If budget or deadline make the model infeasible, a two-stage lexicographic solve adds a non-negative slack to each, first minimizes the slacks (each weighted by $10^9/L$ for its own unit), then fixes them and optimizes the normal objective.
Each stage has its own deterministic-time limit.
This runs only when the solver proves infeasibility, not when it times out.
The families with positive slack are reported as the bottleneck.
We tried CP-SAT's assumption cores for this, but they returned non-minimal sets that named irrelevant literals.

\paragraph{Solver settings.}
CP-SAT runs with one worker, a fixed seed and a limit of 0.5 units of deterministic time, CP-SAT's machine-independent measure of search effort, instead of a wall-clock limit.
In our tests the default 8-worker portfolio returned different, equally good plans from run to run, which breaks reproducibility, while interleaved search was deterministic but too slow for re-solving inside a run.
The returned plan is the best found within the limit and is not guaranteed optimal.
In the reporting runs of our method, \NOptimal{} of \NSolves{} solves (\OptimalPct\%) were proven optimal, \NFeasible{} stopped at the limit with a feasible plan, and \NRelaxed{} went through the budget relaxation.

\section{Rolling-horizon execution}
\label{sec:rolling}

\begin{figure}[t]
\centering
\begin{tikzpicture}[
  node distance=12mm, font=\footnotesize,
  box/.style={draw, rounded corners=2pt, align=center, text width=27mm, minimum height=13mm, inner sep=3pt},
  arr/.style={-{Stealth[length=2mm]}, thick}]
\node[box, fill=blue!6] (plan) {\textbf{Planner}\\template, random DAG or LLM};
\node[box, fill=orange!12, right=of plan] (cp) {\textbf{CP-SAT model}\\skills, context, concurrency, TPM/RPM windows, budget};
\node[box, fill=green!8, right=of cp] (eng) {\textbf{Rolling-horizon engine}\\virtual or real clock, event log};
\node[box, fill=blue!6, right=of eng] (ag) {\textbf{Agent fleet}\\simulator or LiteLLM};
\draw[arr] (plan) -- node[above]{DAG} (cp);
\draw[arr] (cp) -- node[above]{plan} (eng);
\draw[arr] ([yshift=2mm]eng.east) -- node[above]{dispatch} ([yshift=2mm]ag.west);
\draw[arr] (ag.south) to[bend left=28] node[below, align=center]{tokens, latency, failures,\\429s, spawned tasks} (eng.south);
\draw[arr, dashed] (eng.north) to[bend right=28] node[above, align=center]{re-solve on deviation $>25\%$, failure or spawn;\\running tasks pinned, previous plan as hint} (cp.north);
\end{tikzpicture}
\caption{System overview. The engine follows the current plan and calls the solver again whenever execution departs from it.}
\label{fig:arch}
\end{figure}
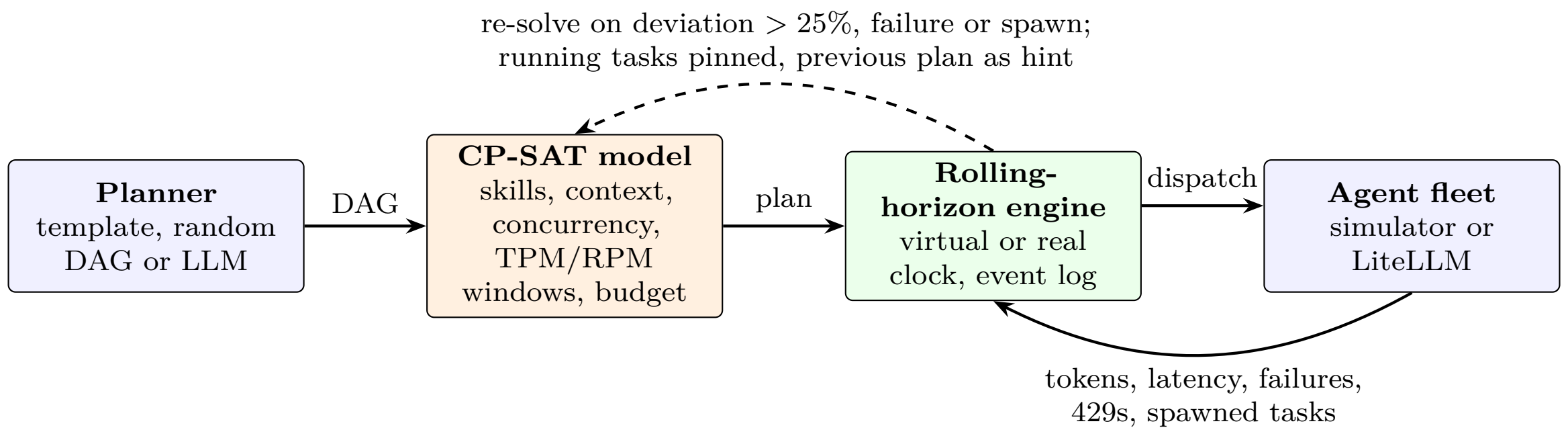

Figure~\ref{fig:arch} shows the loop.
A discrete-event engine dispatches tasks, receives results and decides when to plan again.
With the simulator it runs on a virtual clock, so a 40-task workflow takes milliseconds of wall time; with the LiteLLM runner~\citep{litellm} it awaits real API calls.

\paragraph{Dispatch.}
The engine releases a ready task when its planned agent has a free slot and a shadow copy of the provider's rate window admits the task's estimated tokens.
Tasks are taken in order of planned start.
The plan's start times are therefore targets, not timers: a task can start early if its parents finish early and capacity allows.

\paragraph{Re-solving.}
The engine re-solves the remaining work when a finished task's duration misses its estimate by more than 25\%, when an attempt fails, or when a task spawns children.
Running tasks are pinned to their agents as fixed intervals, recent requests enter as carried windows $R_p$, the estimated cost of running work is reserved from the budget, and the previous plan provides hints: $x_{at}$ for every task it covers, and $s_t$ where its planned start is not in the past.
A task waiting out a 429 back-off is not running, so it is re-planned like any pending task.
Pinning it instead made re-solves infeasible in 3 of 1{,}204 cases in our first reporting run, because its slot had already gone to another task; we fixed this before the results reported here.
Events that happen at the same instant are all applied before the next dispatch, so a re-solve triggered by the first of them still sees the others' tasks as running.
If a re-solve fails, the engine keeps dispatching from the last good plan, and sends ready tasks that plan does not cover, such as children spawned since, to the cheapest eligible agent with a free slot.

\paragraph{Failures and 429s.}
A failed attempt is retried up to three attempts in total, and a task that fails for good skips its descendants.
A 429 triggers exponential back-off of $\min\!\big(60,\ 2\cdot 2^j(1+0.25u)\big)$ seconds, where $j=0,1,\dots$ counts consecutive 429s and $u\in[0,1)$ is a deterministic jitter.
A 429 does not count as an attempt, so the random draw for the next attempt stays the same across strategies.
The count resets on any other outcome, and after 12 consecutive 429s the next one counts as a failed attempt.

\section{Experimental setup}
\label{sec:setup}

\paragraph{Simulator and workflows.}
All strategies are scored against the same seeded simulator.
Realized tokens and durations are the estimates times log-normal noise.
A request that exceeds the context window fails unbilled, a request that the provider's window (on realized tokens) or the agent's concurrency cannot admit gets a 429, and an admitted attempt fails with probability $1-r_{at}$ and is billed.
A successful research task spawns two fact-check tasks with probability 0.3.
Every draw is keyed on (seed, task, attempt) through BLAKE2b and never on the agent or the strategy, so all strategies face the same token counts, failures and spawns; these common random numbers make the comparison paired~\citep{glasserman1992crn}.
A seeded generator builds layered DAGs of 15 to 40 tasks, each needing one of six skills, with log-uniform token estimates.
The budget is 1.25 times the cost of assigning every task of the initial DAG to its cheapest paid eligible agent, so spawned tasks are not budgeted.
Appendix~\ref{app:sim} gives the full details.

\paragraph{Fleet.}
Five agents in three provider pools (Table~\ref{tab:fleet}).
The three Anthropic models share one pool of \TpmAnthropic{} TPM and \RpmAnthropic{} RPM, the OpenAI model has \TpmOpenai{} TPM and \RpmOpenai{} RPM, and the local model is limited only by its single slot.
Anthropic sets limits per model, so this shared pool is stricter than Anthropic's real limits.
These limits and the concurrency caps are set low enough to bind on these workflows.
Prices and per-skill reliabilities are illustrative values, not measurements.
Latency and decoding speed of the four cloud models were measured on \CalDate{} from real API calls.
After one discarded warm-up call per model, we made two calls at each of three output lengths (\CalProbes{} tokens) and fit duration $=$ latency $+$ tokens$/$speed by least squares (\CalSamples{} calls per model, fit RMSE at most \CalRmseMax\,s).
The calls used the thinking settings that workflow steps use: thinking off where the model allows it, otherwise at its lowest level.
The local model's speeds are set by hand.

\begin{table}[t]
\centering
\caption{Benchmark fleet. Conc.\ is the concurrency cap and context is in thousands of tokens. Latency and decoding speed set the duration estimates $d_{at}$; for the cloud models they are measured, with the fit's RMSE shown. Prices are in USD per 1{,}000 tokens.}
\label{tab:fleet}
\footnotesize
\setlength{\tabcolsep}{4pt}
\begin{tabular}{llrrrrrr}
\toprule
Agent & Model & Context & Conc. & Latency (s) & tok/s & RMSE (s) & \$/1k in / out \\
\midrule
\code{opus} & \code{claude-opus-5-5} & 200k & 2 & 0.46 & 117 & 0.176 & 0.004 / 0.02 \\
\code{sonnet} & \code{claude-sonnet-5-5} & 200k & 3 & 0.77 & 154 & 0.323 & 0.002 / 0.01 \\
\code{haiku} & \code{claude-haiku-5-5} & 200k & 4 & 0.41 & 222 & 0.035 & 0.0001 / 0.0005 \\
\code{gpt-mini} & \code{gpt-4o-mini} & 128k & 3 & 0.31 & 67 & 0.104 & 0.00015 / 0.0006 \\
\code{llama-local} & \code{ollama/llama3} & 8k & 1 & 0.30 & 25 & (by hand) & 0 / 0 \\
\bottomrule
\end{tabular}

\end{table}

\paragraph{Strategies.}
\begin{itemize}[leftmargin=*,itemsep=2pt]
  \item \textbf{CP-SAT rolling} (ours), as described above.
  \item \textbf{Online HEFT}, adapted from HEFT~\citep{topcuoglu2002heft}. It ranks ready tasks by upward rank and gives each the agent with the earliest estimated finish among those that can take it now, with the same eligibility filter, shadow rate window, concurrency caps and running-cost budget reserve as our method. It has no look-ahead and no global cost trade-off; Appendix~\ref{app:sim} lists the remaining differences.
  \item \textbf{Greedy cheapest} sends each ready task to the cheapest eligible agent with a free slot. It ignores rate windows and the budget.
  \item \textbf{Reliability router} maps each task once to its most reliable skill-matching agent, in the spirit of static quality-first routing. It ignores capacity, context windows, reliability floors and the budget, and retries failures on the same agent.
  \item \textbf{CP-SAT static} (ablation) solves the same model once at time 0 and dispatches by that plan with the same shadow-window gate. It never re-solves: failed attempts retry on the planned agent, and tasks outside the plan, such as spawned children, go to the cheapest eligible agent with a free slot, without checking the budget.
  \item \textbf{CP-SAT rolling, $w_R=0$} (post-hoc sensitivity) is our method with weights $(0.5,0.5,0)$.
\end{itemize}

\paragraph{Protocol.}
Each strategy runs on DAG seeds 0--19 with three realizations each, \NRuns{} runs per strategy.
All settings were fixed before the first reporting run and not changed afterwards.
The objective weights were tuned on DAG seeds 1000--1019; the rate limits, concurrency caps, budget rule and simulator parameters were set by hand and only checked on those seeds to make each limit bind.
Tuning moved the weights from $(0.5,0.3,0.2)$ to $(0.45,0.45,0.1)$: with lower-bound normalization the risk term's small bound let it dominate and pushed plans toward slow, expensive agents.
Dropping the risk term gave lower cost and fewer budget violations on the tuning seeds but a lower success rate; we kept a small risk weight because the design called for a risk term.
This tuning used the initial, hand-set model speeds, before the calibration described above, so it does not cover the fleet that produced the results.

The reporting seeds were run several times while the system was being finished, after bug fixes and model-speed measurements that were not chosen from benchmark numbers; Appendix~\ref{app:history} lists every run.
Rolling had a lower mean makespan than HEFT in every run, and static beat rolling in all four runs that included it.
Three analyses were chosen after seeing reporting-seed numbers: the DAG-level bootstrap, the budget-relaxation analysis of Section~\ref{sec:results} (prompted by static beating rolling), and the $w_R=0$ check.
No parameter or default was changed in response to any of them.

\paragraph{Metrics and statistics.}
Makespan is the time of the last finish or final failure.
Cost includes failed attempts.
Success rate is finished tasks over initial plus spawned tasks.
A run violates the budget if it spends more than $B$.
Because realizations of one DAG are not independent, we average them per DAG and compute paired differences over the \NDags{} DAGs, with 95\% percentile bootstrap intervals from 10{,}000 resamples~\citep{efron1993bootstrap,field2007bootstrapping}.
On seeds 0--19 we compute \NIntervals{} intervals (nine metrics for each of five comparisons, six metrics shown in the tables, plus two for charged solver time) without a multiplicity correction, so they are descriptive.
Solver wall time is not charged to the simulated clock (Section~\ref{sec:results}).

\paragraph{Reproducibility.}
With one solver worker, a deterministic-time limit and BLAKE2b-derived seeds, the benchmark JSON is byte-identical across reruns and across Python hash seeds on our machine (Windows 11, Python 3.14, OR-Tools 9.15.6755), and a test checks this across processes on a small subset (two tuning DAGs, two realizations, the main strategies).
The data in this paper come from release \code{\SourceRef} (commit \code{\SourceCommit}), except those of Section~\ref{sec:followup} and the fresh-seed figures in Section~\ref{sec:results}, which come from release \code{\FuRef} (commit \code{\FuCommit}); the solver times of the later rerun, also used in its charged gaps, come from commit \code{\FuRerunCommit}.

\section{Results}
\label{sec:results}

\begin{table}[t]
\centering
\caption{Mean per run over \NRuns{} runs (\NDags{} DAGs $\times$ \NReal{} realizations). Bold marks the best of the four main strategies in each column. Budget violations are the fraction of runs that overspent.}
\label{tab:main}
\small
\begin{tabular}{lrrrrrr}
\toprule
Strategy & Makespan (s) & Cost (\$) & \$/task & 429s & Budget viol. & Success \\
\midrule
CP-SAT rolling & \textbf{123.2} & \textbf{0.0798} & \textbf{0.00276} & 1.32 & \textbf{0.38} & 0.974 \\
Online HEFT & 221.9 & 0.0900 & 0.00312 & \textbf{0.78} & 0.73 & 0.970 \\
Greedy cheapest & 167.7 & 0.0808 & 0.00282 & 10.53 & 0.42 & 0.976 \\
Reliability router & 183.8 & 0.7158 & 0.02423 & 38.35 & 1.00 & \textbf{1.000} \\
\midrule \multicolumn{7}{l}{\emph{Ablations}} \\
CP-SAT static & 98.5 & 0.0817 & 0.00285 & 1.53 & 0.38 & 0.964 \\
CP-SAT rolling, $w_R{=}0$ & 113.6 & 0.0768 & 0.00268 & 1.18 & 0.27 & 0.973 \\
\bottomrule
\end{tabular}

\end{table}

\begin{table}[t]
\centering
\caption{Paired differences against CP-SAT rolling (row strategy minus rolling): mean [95\% CI], bootstrapped over the \NDags{} DAGs. Positive means the other strategy is higher. Shaded cells exclude zero.}
\label{tab:ci}
\footnotesize
\setlength{\tabcolsep}{3pt}
\begin{tabular}{@{}c@{}}
\textbf{(a) Baselines}\\[2pt]
\begin{tabular}{lrrr}
\toprule
Metric & Online HEFT & Greedy cheapest & Reliability router \\
\midrule
Makespan (s) & \cellcolor{sig}98.8 [59.9, 140.6] & \cellcolor{sig}44.6 [22.8, 71.7] & \cellcolor{sig}60.6 [35.5, 85.0] \\
Cost (\$) & \cellcolor{sig}0.0103 [0.0073, 0.0133] & 0.0011 [$-$0.0036, 0.0064] & \cellcolor{sig}0.6360 [0.5525, 0.7195] \\
\$/task & \cellcolor{sig}0.00036 [0.00026, 0.00045] & 0.00006 [$-$0.00012, 0.00029] & \cellcolor{sig}0.02148 [0.02003, 0.02294] \\
429s & \cellcolor{sig}$-$0.53 [$-$1.05, $-$0.10] & \cellcolor{sig}9.22 [4.48, 14.73] & \cellcolor{sig}37.03 [30.50, 43.90] \\
Budget viol. & \cellcolor{sig}0.35 [0.20, 0.50] & 0.03 [$-$0.10, 0.18] & \cellcolor{sig}0.62 [0.52, 0.72] \\
Success & $-$0.004 [$-$0.010, 0.000] & 0.001 [0.000, 0.003] & \cellcolor{sig}0.026 [0.005, 0.054] \\
\bottomrule
\end{tabular}
\\[8pt]
\textbf{(b) Ablations}\\[2pt]
\begin{tabular}{lrr}
\toprule
Metric & CP-SAT static & CP-SAT rolling, $w_R{=}0$ \\
\midrule
Makespan (s) & \cellcolor{sig}$-$24.6 [$-$42.6, $-$8.8] & \cellcolor{sig}$-$9.6 [$-$19.2, $-$1.3] \\
Cost (\$) & 0.0019 [$-$0.0020, 0.0058] & \cellcolor{sig}$-$0.0030 [$-$0.0048, $-$0.0015] \\
\$/task & 0.00009 [$-$0.00001, 0.00022] & \cellcolor{sig}$-$0.00008 [$-$0.00013, $-$0.00004] \\
429s & 0.22 [$-$0.22, 0.63] & $-$0.13 [$-$0.55, 0.23] \\
Budget viol. & 0.00 [$-$0.08, 0.10] & \cellcolor{sig}$-$0.12 [$-$0.20, $-$0.03] \\
Success & $-$0.011 [$-$0.032, 0.001] & $-$0.002 [$-$0.005, 0.001] \\
\bottomrule
\end{tabular}

\end{tabular}
\end{table}

\begin{figure}[t]
\centering
\includegraphics[width=\linewidth]{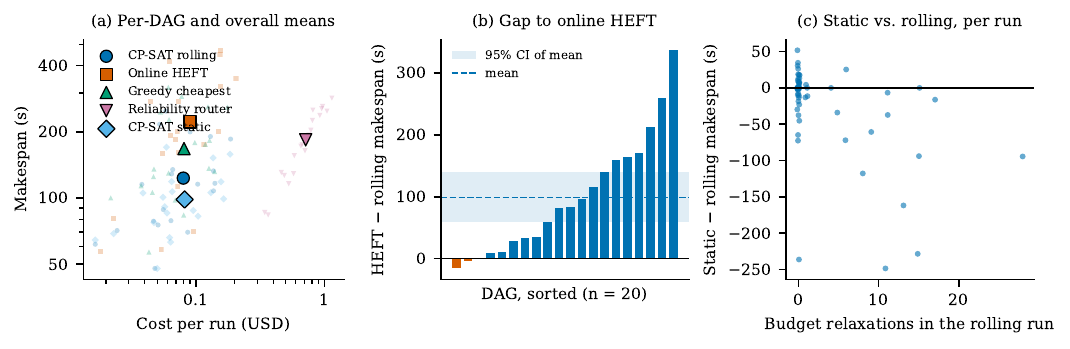}
\caption{(a) Cost and makespan per DAG (small markers) and overall means (large markers), log scales. (b) Per-DAG makespan of online HEFT minus CP-SAT rolling, sorted, with the mean and its 95\% CI; blue bars are DAGs where rolling finishes first, gray bars ties within \TiePct\%. (c) Per run, makespan of CP-SAT static minus CP-SAT rolling against the number of budget relaxations in the rolling run (horizontal jitter added); points below zero are runs where static finishes first.}
\label{fig:results}
\end{figure}

Table~\ref{tab:main} gives the means and Table~\ref{tab:ci} the paired differences.

\paragraph{Planning beats the list scheduler.}
Online HEFT is the strongest heuristic baseline.
CP-SAT rolling finishes \MkDiffHeft\,s sooner (95\% CI [\MkDiffHeftLo, \MkDiffHeftHi]), a \MkGainPct\% reduction, and costs \CostGainPct\% less.
It finishes first on \MkWinsHeft{} of the \NDags{} DAGs, HEFT on \MkLossHeft, with \MkTieHeft{} within \TiePct\% (Figure~\ref{fig:results}b), and it is cheaper on \CostWinsHeft.
HEFT's budget-violation rate is \BvDiffHeft{} higher (CI [\BvDiffHeftLo, \BvDiffHeftHi]), and we find no difference in success rate.
Restricted to the \RestrictedHeftN{} runs in which both complete every task, rolling still finishes \RestrictedHeft\,s sooner on average.
HEFT draws \RlHeftFewer{} fewer 429s per run (CI [\RlHeftFewerLo, \RlHeftFewerHi]).
Our reading of the gap is that HEFT commits each ready task to the agent that finishes it soonest under current conditions, without seeing that the choice uses up a fast agent's slots, the provider's window or the budget that later tasks need.

\paragraph{Other baselines.}
Against greedy cheapest, we find no difference in cost or success rate, and CP-SAT rolling is \MkDiffCheap\,s faster and draws \RlDiffCheap{} fewer 429s per run, because greedy cheapest sends requests to the cheapest provider without tracking its window.
The reliability router has the highest success rate, \SuccRouter{} against \SuccOurs{} (difference \SuccDiffRouter, CI [\SuccDiffRouterLo, \SuccDiffRouterHi]).
It always uses the most reliable model, which costs about \RouterCostX$\times$ as much, takes \RouterMkX$\times$ as long and draws about \RouterRl{} 429s per run, since it ignores capacity.
Which of these points is better depends on what a failed task costs the user, and our objective weights encode only one choice.

\paragraph{Re-planning can hurt.}
The static ablation solves the same model once and never re-plans.
It finishes \StaticFaster\,s sooner than rolling (CI [\StaticFasterLo, \StaticFasterHi]) and first on \StaticWins{} of \NDags{} DAGs, with no detectable difference in cost, budget violations or success rate (Table~\ref{tab:ci}b); restricted to the \RestrictedStaticN{} runs in which both complete every task, it is \RestrictedStatic\,s faster.
It has the lowest mean makespan of all strategies, even though about \StaticUnplanned{} spawned tasks per run fall outside its plan and go to a budget-blind fallback, which also helps explain why its budget record is no worse.

A post-hoc analysis, prompted by this result, points to how rolling handles the budget.
Realized spending drifts above the estimates: failed attempts are billed, token counts vary, and spawned tasks are not in the budget.
Once the remaining budget, after reserving the estimated cost of running tasks, is below the cost of the cheapest assignment of the pending tasks, a re-solve is infeasible.
The relaxation then minimizes the overspend first and optimizes the objective only among plans with that minimal overspend, which puts tasks on their cheapest eligible agents.
For writing, fact-check and summary tasks that is often the free local model, the slowest agent and limited to one task at a time; research tasks are unaffected, because their cheapest agent is also the fastest.
Static never observes spending, so it never makes this move.
\PanicRunsWith{} of the \NRuns{} rolling runs, from \NRelaxedDags{} DAGs, relaxed the budget at least once.
In those runs static was \PanicGapWith\,s faster on average, against \PanicGapWithout\,s in the other \PanicRunsWithout.
Re-running them, the local model's share of rolling's dispatches rose from \LlamaBefore\% before the first relaxation to \LlamaAfter\% after it, against \LlamaStatic\% for static on the same tasks.
The effect is concentrated: four runs account for \TopFourShare\% of the gap in the relaxed runs, and across DAGs the correlation between mean relaxations and the static-minus-rolling gap is $r$\,=\,\DagR{} with a 95\% bootstrap interval of [\DagRLo, \DagRHi] that includes zero; Figure~\ref{fig:results}c shows the per-run pattern.
On these seeds alone, the account is plausible but not established.
Any fix suggested by these seeds needs fresh seeds to evaluate, so Section~\ref{sec:followup} tests two on new workflows.

\paragraph{Risk weight.}
As a post-hoc check on the reporting seeds, we reran rolling without the risk term ($w_R=0$).
It is \RiskZeroFaster\,s faster (CI [\RiskZeroFasterLo, \RiskZeroFasterHi]), also cheaper with fewer budget violations, with no detectable difference in success rate (Table~\ref{tab:ci}b).
On these seeds, the small risk weight we kept after tuning costs time and money without a measurable gain in completed tasks.
That does not hold on fresh seeds, where dropping the risk term lowers the success rate (Section~\ref{sec:followup}).
We report the tuned configuration as the main result and this one as a sensitivity check.

\paragraph{Limits are not guarantees.}
CP-SAT rolling still draws \RlOurs{} 429s per run and overspends in \BvOurs{} of its runs.
Its plans are feasible on estimated tokens, but realized tokens drift, and spawned tasks are not covered by the budget.

\paragraph{Solver time.}
Rolling solves about \SolvesOurs{} times per run: once at the start and about \ResolvesOurs{} re-solves.
Over \NSolves{} solves, the median wall time is \SolveMsPFifty\,ms, the 95th percentile \SolveMsPNinetyFive\,ms, the 99th \SolveMsPNinetyNine\,ms and the maximum \SolveMsMax\,ms, for a mean of \SolveSRunMean\,s per run (maximum \SolveSRunMax\,s).
These are wall-clock measurements on Windows 11 with Python 3.14 and OR-Tools 9.15, and unlike the benchmark metrics they vary between runs.
The benchmark does not charge solver time to the simulated clock.
Charging it, as if every solve blocked all dispatching, leaves rolling \ChargedHeft\,s ahead of HEFT (CI [\ChargedHeftLo, \ChargedHeftHi]) and widens static's lead over rolling to \ChargedStatic\,s (CI [\ChargedStaticLo, \ChargedStaticHi]).
These figures are for seeds 0--19 (\NRuns{} runs).
Totals reach higher in the \FuRuns{} runs on the fresh seeds of Section~\ref{sec:followup}, which were timed separately.
There, rolling's solves took \FuSolveMean\,s per run on average in the original benchmark and \FuSolveRerunMean\,s in a later timing rerun.
The same run, with a makespan of \FuSolveWorstMk\,s, had the largest total both times: \FuSolveMax\,s in the original and \FuSolveRerunMax\,s in the rerun.
Solving took more than 10\,s in \FuSolveOverTen{} of the \FuRuns{} runs in the original and \FuSolveRerunOverTen{} in the rerun.
Another of our benchmarks shared the CPU for part of the original run but not the rerun, yet the rerun was slower; we did not control for other load, power or temperature and do not know why, so both are rough measurements.
Charging solver time in the same way on the fresh seeds, an analysis we did not pre-register, leaves rolling \FuSolveChargedHeft\,s ahead of HEFT with the original times (CI [\FuSolveChargedHeftLo, \FuSolveChargedHeftHi]) and \FuSolveRerunChargedHeft\,s with the rerun's (CI [\FuSolveRerunChargedHeftLo, \FuSolveRerunChargedHeftHi]), against \FuMkHeft\,s uncharged.
Static's lead over rolling, \FuStaticFaster\,s uncharged, becomes \FuSolveChargedStatic\,s with the original times (CI [\FuSolveChargedStaticLo, \FuSolveChargedStaticHi]) and \FuSolveRerunChargedStatic\,s with the rerun's (CI [\FuSolveRerunChargedStaticLo, \FuSolveRerunChargedStaticHi]).
These four intervals are uncorrected and descriptive.

\paragraph{Live execution.}
To check that the system runs end to end on real APIs, we ran a hand-written 11-task research-report workflow once with CP-SAT rolling through LiteLLM, on a fleet of the four cloud models of Table~\ref{tab:fleet} with looser rate limits and concurrency caps than the benchmark uses; the plan assigned tasks to \LiveModelsUsed{} of them.
Appendix~\ref{app:live} gives the prompt and token settings.
All \LiveDone{} of \LiveTasks{} tasks succeeded in \LiveMakespan\,s for \$\LiveCost, with \LiveRateLimited{} 429s, \LiveTruncated{} truncated outputs and \LiveReasoning{} reasoning tokens.
The models wrote \LiveRatioMin{} to \LiveRatioMax{} times the template's output estimates, so most tasks ran longer than planned and the engine re-solved \LiveReSolves{} times after its initial solve.
This single run shows that the pieces work together with real APIs; it is not evidence of performance.

\section{Pre-registered follow-up on fresh seeds}
\label{sec:followup}

Release \code{\FuRef} (commit \code{\FuCommit}) tests the budget account of Section~\ref{sec:results} on workflows that played no part in forming it.
The selection and promotion rules below were committed to the repository (commit \code{\PreregCommit}) before the parameter sweep and before any run on the fresh seeds, and the file was not edited afterwards.

\paragraph{Two opt-in changes.}
Both are solver options that are off by default.
With them added to the code, seeds 0--19 were re-run, and every result behind Tables~\ref{tab:main} and~\ref{tab:ci} came out byte-identical.
\emph{Priced overspend} replaces the lexicographic relaxation of Section~\ref{sec:model}.
When the budget or deadline cannot be met, a single solve allows slack and charges each unit $\lambda$ times the matching objective coefficient, on top of the objective's own cost and makespan terms, so an overspent micro-dollar weighs $1+\lambda$ times a budgeted one.
Solves in which the budget can be met are unchanged.
\emph{Paid reference} floors the cost term's lower bound $L_C$ at the cheapest plan over paid agents.
Without it, $L_C$ falls to one micro-dollar once only free work remains, and cost then outweighs makespan.
The paid reference would become the default only if, against rolling, its makespan interval lay below zero, its overrun interval's upper bound were at most \$0.0005 and its mean success rate fell by no more than 0.005.
Priced overspend spends more by design, so it stays opt-in and is reported as a trade-off whatever it shows.

\paragraph{Choosing $\lambda$.}
On the tuning seeds 1000--1019, the rule took, among $\lambda\in\{0,0.25,0.5,1,2,4,8\}$ with mean overrun at most static's, the one with the lowest mean makespan, counting makespans within 0.5\% as tied and breaking ties toward the larger $\lambda$.
All \SweepQualified{} values qualified and $\lambda=$ \SweepTied{} tied, so $\lambda=\SweepLambda$ was used.
Since even $\lambda=0$ tied, the speed-up comes mainly from trading overspend against time in one solve rather than minimizing it first; the larger values 4 and 8 were slower on these 20 DAGs.

\paragraph{Protocol.}
One benchmark run, which the pre-registration allows only once, covered \FuDags{} fresh DAGs (seeds 2000--2079) with \FuReal{} realizations each, \FuRuns{} paired runs per strategy, and every strategy and ablation.
We later re-executed that benchmark once, on unchanged code and only to time the solver (Section~\ref{sec:results}).
Its results file was byte-identical, so it is not a second test, and nothing was re-tuned.
Statistics are as in Section~\ref{sec:setup}.
Only the comparisons covered by the pre-registered rules are confirmatory; the table's other \FuIntervals{} intervals, and every contrast below that we call not pre-registered, are uncorrected and descriptive.

\begin{table}[t]
\centering
\caption{Pre-registered follow-up on \FuDags{} fresh DAGs (seeds 2000--2079, \FuReal{} realizations each, release \code{\FuRef}): paired differences against CP-SAT rolling (row minus rolling), mean [95\% CI] over DAGs. Rolling's means are \FuMkRoll\,s and \$\FuCostRoll{} per run. Overrun is spending above the budget per run. Shaded cells exclude zero, judged on unrounded bounds. The priced and paid-reference rows test the pre-registered changes; the other rows repeat Tables~\ref{tab:main} and~\ref{tab:ci} on new workflows.}
\label{tab:followup}
\footnotesize
\setlength{\tabcolsep}{3pt}
\begin{tabular}{lrrrr}
\toprule
Strategy & Makespan (s) & Cost (\$) & Overrun (\$) & Success \\
\midrule
Online HEFT & \cellcolor{sig}142.4 [117.6, 168.0] & \cellcolor{sig}0.0132 [0.0111, 0.0155] & \cellcolor{sig}0.0068 [0.0057, 0.0080] & $-$0.007 [$-$0.019, 0.005] \\
Greedy cheapest & \cellcolor{sig}42.6 [29.6, 56.2] & $-$0.0016 [$-$0.0050, 0.0016] & 0.0013 [$-$0.0005, 0.0030] & $-$0.011 [$-$0.026, 0.003] \\
Reliability router & \cellcolor{sig}74.5 [62.1, 86.4] & \cellcolor{sig}0.6843 [0.6398, 0.7275] & \cellcolor{sig}0.6766 [0.6326, 0.7198] & \cellcolor{sig}0.021 [0.011, 0.032] \\
\midrule \multicolumn{5}{l}{\emph{CP-SAT variants}} \\
Static & \cellcolor{sig}$-$21.2 [$-$31.0, $-$12.5] & 0.0009 [$-$0.0005, 0.0024] & \cellcolor{sig}0.0010 [0.0003, 0.0017] & $-$0.010 [$-$0.021, 0.000] \\
Priced ($\lambda{=}2$) & \cellcolor{sig}$-$15.2 [$-$23.2, $-$8.5] & \cellcolor{sig}0.0005 [0.0003, 0.0007] & \cellcolor{sig}0.0004 [0.0003, 0.0006] & \cellcolor{sig}0.003 [0.000, 0.009] \\
Paid reference & $-$1.4 [$-$4.0, 1.6] & \cellcolor{sig}0.0003 [0.0001, 0.0006] & \cellcolor{sig}0.0001 [0.0000, 0.0002] & 0.000 [$-$0.001, 0.000] \\
$w_R{=}0$ & $-$3.5 [$-$8.5, 1.7] & \cellcolor{sig}$-$0.0045 [$-$0.0059, $-$0.0032] & \cellcolor{sig}$-$0.0013 [$-$0.0018, $-$0.0008] & \cellcolor{sig}$-$0.010 [$-$0.019, $-$0.003] \\
\bottomrule
\end{tabular}

\end{table}

\paragraph{Priced overspend.}
Priced overspend finishes \FuPricedFaster\,s sooner than rolling (CI [\FuPricedFasterLo, \FuPricedFasterHi]); it is faster on \FuPricedBetter{} DAGs, slower on \FuPricedWorse{} and tied on \FuPricedTie{} (Table~\ref{tab:followup}).
It overspends by \$\FuPricedOver{} more per run (CI [\FuPricedOverLo, \FuPricedOverHi]) and its success rate is \FuPricedSuccPts{} points higher (CI [\FuPricedSuccPtsLo, \FuPricedSuccPtsHi]).
On these seeds static leads rolling by \FuStaticFaster\,s (CI [\FuStaticFasterLo, \FuStaticFasterHi]), so priced closes about \FuPricedShare\% of that lead (a ratio of mean differences; DAG bootstrap CI [\FuPricedShareLo, \FuPricedShareHi]).
Priced differs from rolling only in solves where the budget is already infeasible, so this is an intervention on the proposed mechanism, on seeds that did not suggest it, and it supports the account.
It does not explain the whole gap.
In a comparison we did not pre-register, static is still \PhStaticFaster\,s faster than priced (CI [\PhStaticFasterLo, \PhStaticFasterHi]), with no detectable difference in overrun, while priced completes \PhSuccPts{} points more tasks (CI [\PhSuccPtsLo, \PhSuccPtsHi]).
The correlational evidence is also clearer than on seeds 0--19.
Of the \FuRuns{} rolling runs, \FuPanicWith{} relaxed the budget at least once; in those runs static was \FuPanicGapWith\,s faster on average, against \FuPanicGapWithout\,s in the other \FuPanicWithout{}, and across DAGs the correlation between mean relaxations and the static-minus-rolling gap is $r$\,=\,\FuDagR{} (CI [\FuDagRLo, \FuDagRHi]).

\paragraph{Paid reference: a null result.}
The paid reference fails its promotion rule: its makespan difference is \FuPaidMk\,s (CI [\FuPaidMkLo, \FuPaidMkHi]), so it stays an ablation and the default is unchanged.
On seeds 0--19, which produced the budget hypothesis and are not evidence for it, the same comparison gave \OldPaidMk\,s (CI [\OldPaidMkLo, \OldPaidMkHi]); judged on those seeds alone, it would have shipped.

\paragraph{Replication and one change.}
The main result replicates.
Online HEFT is \FuMkHeft\,s slower than rolling (CI [\FuMkHeftLo, \FuMkHeftHi]) and costs \$\FuCostHeft{} more per run (CI [\FuCostHeftLo, \FuCostHeftHi]), and rolling finishes first on \FuWinsHeft{} of the \FuDags{} DAGs.
Greedy cheapest is \FuMkCheap\,s and the reliability router \FuMkRouter\,s slower than rolling.
Static's lead over rolling also replicates, and with four times as many DAGs its extra overspend is now detectable, at \$\FuStaticOver{} per run (CI [\FuStaticOverLo, \FuStaticOverHi]).
One sensitivity result changes.
Without the risk term, the success rate changes by \FuRiskZeroSuccPts{} points (CI [\FuRiskZeroSuccPtsLo, \FuRiskZeroSuccPtsHi]), with no detectable difference in makespan and a lower cost.
On seeds 0--19 this loss was too small to detect, so the conclusion drawn there in Section~\ref{sec:results}, that the risk weight buys no completed tasks, does not hold.

\section{Related work}
\label{sec:related}

\paragraph{Model routing and cascades.}
FrugalGPT, RouteLLM, Hybrid LLM and AutoMix pick a model, or a sequence of models, for each query to trade answer quality against price~\citep{chen2023frugalgpt,ong2024routellm,ding2024hybridllm,aggarwal2023automix}, and RouterBench compares such routers~\citep{hu2024routerbench}.
They decide per query and do not model task dependencies or capacity shared between queries.
Our reliability router is a simple stand-in for quality-first routing; a learned router would sit in the same place in the loop and could supply the reliabilities $r_{a,s}$ our model consumes.

\paragraph{Agent and workflow frameworks.}
AutoGen, MetaGPT, HuggingGPT and LLMCompiler~\citep{wu2023autogen,hong2023metagpt,shen2023hugginggpt,kim2023llmcompiler} express multi-step LLM work as conversations, role pipelines or task graphs, and LLMCompiler runs independent calls of a graph in parallel.
Our planner produces the same kind of graph; the contribution here is in deciding where and when its nodes run.

\paragraph{Scheduling LLM workflows.}
Serving systems such as Parrot~\citep{lin2024parrot} and Autellix~\citep{luo2025autellix} schedule the calls of multi-call programs inside the inference engine to reduce latency.
FATE~\citep{huang2026fate} is closest in technique: it re-plans over the ready frontier of a heterogeneous LLM workflow with a CP-SAT-backed planner, but it works on the serving side, placing work on GPUs, and its abstract mentions neither provider rate limits nor a dollar budget.
On the client side, Wang et al.~\citep{wang2026ontime} allocate models to a dependency-structured workflow online under budget and deadline constraints, BAMAS~\citep{yang2025bamas} selects LLMs under a budget with an integer program before execution, and Kosorin et al.~\citep{kosorin2026dispatch} and HiveMind~\citep{agyemang2026hivemind} gate dispatch against external request and token limits with heuristic admission control.
Our work combines the three elements these systems treat separately: model choice with a budget, provider limits encoded as constraints rather than enforced by a gate, and re-planning during execution.
Our characterization of the 2025--2026 systems rests on their published abstracts.

\paragraph{Scheduling theory.}
The model is a multi-mode resource-constrained project scheduling problem~\citep{hartmann2010rcpsp} with cumulative resources~\citep{aggoun1993cumulative}, solved with CP-SAT~\citep{ortools,cpsat}.
HEFT~\citep{topcuoglu2002heft} is the standard list-scheduling heuristic for heterogeneous processors, and re-solving with in-progress work fixed is the predictive-reactive approach to dynamic scheduling~\citep{ouelhadj2009dynamic,vieira2003rescheduling}.

\section{Limitations}
\label{sec:limits}

\begin{itemize}[leftmargin=*,itemsep=2pt]
  \item \textbf{Simulated evaluation.} The benchmark runs against a simulator, not real APIs. The simulator applies the same request-time window and concurrency rules that the model encodes, which favors a planner that models those rules. The noise is real: realized tokens differ from the plan, so the planner still draws 429s.
  \item \textbf{Provider metering.} The window encoding and the simulator both count tokens at request time in a 60-second sliding window. This is safe under a single token bucket (Section~\ref{sec:model}) but does not model separate input and output limits, sub-minute enforcement, ramp limits, or OpenAI's use of \code{max\_tokens}. Real 429 rates may therefore differ from the simulated ones.
  \item \textbf{One fleet, one generator.} Results cover one fleet with illustrative prices and reliabilities and one random DAG family. Other fleets, tighter or looser limits, or real workflow shapes may change the ranking.
  \item \textbf{Sample size and metrics.} The main comparison uses twenty DAGs, a small sample with wide intervals; the follow-up uses \FuDags. Nine metrics are compared without correction.
  \item \textbf{Run history.} The reporting seeds produced six sets of results during development (Appendix~\ref{app:history}). No parameter was chosen from those runs, but the budget-relaxation analysis and the $w_R=0$ check were chosen after seeing them. On fresh seeds, the follow-up supports the first and finds a loss in success rate that the second missed.
  \item \textbf{Budget rule.} Priced overspend is faster but spends more, and static remains faster still, so the follow-up supports the budget account without explaining all of static's lead. Priced overspend is not the default.
  \item \textbf{Estimates.} The planner is only as good as its token and duration estimates. In the live run, models wrote \LiveRatioMin{} to \LiveRatioMax{} times the template's output estimates.
  \item \textbf{Pooled limits.} The benchmark pools three models under one rate limit, which real providers that limit per model do not do. The ranking may differ under per-model limits.
  \item \textbf{Solver budget.} One worker with a 0.5 deterministic-time limit trades plan quality for reproducibility. Larger workflows may need a different trade-off.
  \item \textbf{Platform.} The published results file was produced and byte-checked on Windows; continuous integration passes on Linux and Windows, but whether Linux reproduces the same bytes has not been compared.
\end{itemize}

\section{Conclusion}

Treating a multi-model LLM workflow as a scheduling problem, with provider limits encoded as cumulative constraints over request windows, beat a capacity-aware list scheduler by a wide margin in simulation: shorter makespan, lower cost and fewer budget overruns, with no detectable difference in success rate.
Solving took a median of \SolveMsPFifty\,ms per solve and a mean total of \SolveSRunMean\,s per run; on the fresh seeds, that mean was \FuSolveMean\,s in the original benchmark and \FuSolveRerunMean\,s in the timing rerun, and one run's total reached \FuSolveRerunMax\,s.
Re-planning was the weak part.
A single up-front solve was faster still, and a post-hoc analysis traced much of the gap to re-solves that relax an infeasible budget by minimizing the overspend first.
On \FuDags{} fresh workflows, a pre-registered change that prices the overspend instead closed about \FuPricedShare\% of static's lead (CI [\FuPricedShareLo, \FuPricedShareHi]) at a small extra spend, which supports that account, and the advantage over HEFT held.
Static remained faster, in a comparison we did not pre-register, so the account does not explain the whole gap.
The method also does not eliminate 429s or budget overruns, and a quality-first router completes more tasks at many times the cost.
The next steps are budget rules that keep the speed of priced overspend without its extra spend, metering models that match each provider's documentation, and a benchmark on real APIs with real task outputs.

\section*{Code and data availability}
The code and the benchmark results are available at \url{https://github.com/pouriamrt/neuro-orch} (releases \code{\SourceRef} and \code{\FuRef}, plus commit \code{\FuRerunCommit} for the rerun's solver times; MIT license).
The scripts that derive every number, table and figure in this paper from those sources, together with the extracted data, are included as ancillary files with this submission.

\section*{Competing interests and funding}
The authors declare no competing interests. This work received no specific funding.

\bibliographystyle{plainnat}
\bibliography{refs}

\begin{thebibliography}{30}
\providecommand{\natexlab}[1]{#1}
\providecommand{\url}[1]{\texttt{#1}}
\expandafter\ifx\csname urlstyle\endcsname\relax
  \providecommand{\doi}[1]{doi: #1}\else
  \providecommand{\doi}{doi: \begingroup \urlstyle{rm}\Url}\fi

\bibitem[Aggarwal et~al.(2023)Aggarwal, Madaan, Anand,
  et~al.]{aggarwal2023automix}
Pranjal Aggarwal, Aman Madaan, Ankit Anand, et~al.
\newblock {AutoMix}: Automatically mixing language models.
\newblock arXiv:2310.12963, 2023.

\bibitem[Aggoun and Beldiceanu(1993)]{aggoun1993cumulative}
Abderrahmane Aggoun and Nicolas Beldiceanu.
\newblock Extending {CHIP} in order to solve complex scheduling and placement
  problems.
\newblock \emph{Mathematical and Computer Modelling}, 17\penalty0 (7):\penalty0
  57--73, 1993.
\newblock \doi{10.1016/0895-7177(93)90068-A}.

\bibitem[Agyemang et~al.(2026)Agyemang, Kponyo, Somuah,
  et~al.]{agyemang2026hivemind}
Justice~Owusu Agyemang, Jerry~John Kponyo, Obed~Kwasi Somuah, et~al.
\newblock {HiveMind}: {OS}-inspired scheduling for concurrent {LLM} agent
  workloads.
\newblock arXiv:2604.17111, 2026.

\bibitem[{Anthropic}(2026)]{anthropicratelimits}
{Anthropic}.
\newblock Rate limits.
\newblock \url{https://platform.claude.com/docs/en/api/rate-limits}, 2026.
\newblock Accessed 2026-11-29.

\bibitem[{BerriAI}(2026)]{litellm}
{BerriAI}.
\newblock {LiteLLM}.
\newblock \url{https://github.com/BerriAI/litellm}, 2026.
\newblock Accessed 2026-11-29.

\bibitem[Chen et~al.(2023)Chen, Zaharia, and Zou]{chen2023frugalgpt}
Lingjiao Chen, Matei Zaharia, and James Zou.
\newblock {FrugalGPT}: How to use large language models while reducing cost and
  improving performance.
\newblock arXiv:2305.05176, 2023.

\bibitem[Ding et~al.(2024)Ding, Mallick, Wang, Sim, Mukherjee, R{\"u}hle,
  Lakshmanan, and Awadallah]{ding2024hybridllm}
Dujian Ding, Ankur Mallick, Chi Wang, Robert Sim, Subhabrata Mukherjee, Victor
  R{\"u}hle, Laks V.~S. Lakshmanan, and Ahmed~Hassan Awadallah.
\newblock Hybrid {LLM}: Cost-efficient and quality-aware query routing.
\newblock In \emph{International Conference on Learning Representations
  (ICLR)}, 2024.
\newblock arXiv:2404.14618.

\bibitem[Efron and Tibshirani(1993)]{efron1993bootstrap}
Bradley Efron and Robert~J. Tibshirani.
\newblock \emph{An Introduction to the Bootstrap}.
\newblock Chapman \& Hall/CRC, 1993.
\newblock \doi{10.1201/9780429246593}.

\bibitem[Field and Welsh(2007)]{field2007bootstrapping}
Christopher~A. Field and Alan~H. Welsh.
\newblock Bootstrapping clustered data.
\newblock \emph{Journal of the Royal Statistical Society: Series B (Statistical
  Methodology)}, 69\penalty0 (3):\penalty0 369--390, 2007.
\newblock \doi{10.1111/j.1467-9868.2007.00593.x}.

\bibitem[Glasserman and Yao(1992)]{glasserman1992crn}
Paul Glasserman and David~D. Yao.
\newblock Some guidelines and guarantees for common random numbers.
\newblock \emph{Management Science}, 38\penalty0 (6):\penalty0 884--908, 1992.
\newblock \doi{10.1287/mnsc.38.6.884}.

\bibitem[Hartmann and Briskorn(2010)]{hartmann2010rcpsp}
S{\"o}nke Hartmann and Dirk Briskorn.
\newblock A survey of variants and extensions of the resource-constrained
  project scheduling problem.
\newblock \emph{European Journal of Operational Research}, 207\penalty0
  (1):\penalty0 1--14, 2010.
\newblock \doi{10.1016/j.ejor.2009.11.005}.

\bibitem[Hong et~al.(2023)Hong, Zhuge, Chen, et~al.]{hong2023metagpt}
Sirui Hong, Mingchen Zhuge, Jiaqi Chen, et~al.
\newblock {MetaGPT}: Meta programming for a multi-agent collaborative
  framework.
\newblock arXiv:2308.00352, 2023.

\bibitem[Hu et~al.(2024)Hu, Bieker, Li, et~al.]{hu2024routerbench}
Qitian~Jason Hu, Jacob Bieker, Xiuyu Li, et~al.
\newblock {RouterBench}: A benchmark for multi-{LLM} routing system.
\newblock arXiv:2403.12031, 2024.

\bibitem[Huang et~al.(2026)Huang, Hu, Wu, and Li]{huang2026fate}
Zirui Huang, Yi-Xiang Hu, Feng Wu, and Xiangyang Li.
\newblock {FATE}: Future-state-aware scheduling for heterogeneous {LLM}
  workflows.
\newblock arXiv:2605.07238, 2026.

\bibitem[Kim et~al.(2023)Kim, Moon, Tabrizi, Lee, Mahoney, Keutzer, and
  Gholami]{kim2023llmcompiler}
Sehoon Kim, Suhong Moon, Ryan Tabrizi, Nicholas Lee, Michael~W. Mahoney, Kurt
  Keutzer, and Amir Gholami.
\newblock An {LLM} compiler for parallel function calling.
\newblock arXiv:2312.04511, 2023.

\bibitem[Kosorin et~al.(2026)Kosorin, Schirmer, Wossnig, and
  Bermbach]{kosorin2026dispatch}
Stanislav Kosorin, Trever Schirmer, Leonard Wossnig, and David Bermbach.
\newblock Dispatch control for priority-based, multi-tenant workflows under
  multi-resource external rate limits.
\newblock arXiv:2611.02926, 2026.

\bibitem[Lin et~al.(2024)Lin, Han, Zhang, Yang, Yang, Chen, and
  Qiu]{lin2024parrot}
Chaofan Lin, Zhenhua Han, Chengruidong Zhang, Yuqing Yang, Fan Yang, Chen Chen,
  and Lili Qiu.
\newblock Parrot: Efficient serving of {LLM}-based applications with semantic
  variable.
\newblock In \emph{18th USENIX Symposium on Operating Systems Design and
  Implementation (OSDI 24)}, pages 929--945. USENIX Association, 2024.

\bibitem[Luo et~al.(2025)Luo, Shi, Cai, et~al.]{luo2025autellix}
Michael Luo, Xiaoxiang Shi, Colin Cai, et~al.
\newblock Autellix: An efficient serving engine for {LLM} agents as general
  programs.
\newblock arXiv:2502.13965, 2025.

\bibitem[Ong et~al.(2024)Ong, Almahairi, Wu, Chiang, Wu, Gonzalez, Kadous, and
  Stoica]{ong2024routellm}
Isaac Ong, Amjad Almahairi, Vincent Wu, Wei-Lin Chiang, Tianhao Wu, Joseph~E.
  Gonzalez, M.~Waleed Kadous, and Ion Stoica.
\newblock {RouteLLM}: Learning to route {LLMs} with preference data.
\newblock arXiv:2406.18665, 2024.

\bibitem[{OpenAI}(2026)]{openairatelimits}
{OpenAI}.
\newblock Rate limits.
\newblock \url{https://developers.openai.com/api/docs/guides/rate-limits},
  2026.
\newblock Accessed 2026-11-29.

\bibitem[Ouelhadj and Petrovic(2009)]{ouelhadj2009dynamic}
Djamila Ouelhadj and Sanja Petrovic.
\newblock A survey of dynamic scheduling in manufacturing systems.
\newblock \emph{Journal of Scheduling}, 12\penalty0 (4):\penalty0 417--431,
  2009.
\newblock \doi{10.1007/s10951-008-0090-8}.

\bibitem[Perron and Didier(2025)]{cpsat}
Laurent Perron and Fr{\'e}d{\'e}ric Didier.
\newblock {CP-SAT}.
\newblock \url{https://developers.google.com/optimization/cp/cp_solver/}, 2025.

\bibitem[Perron and Furnon(2025)]{ortools}
Laurent Perron and Vincent Furnon.
\newblock {OR-Tools}.
\newblock \url{https://developers.google.com/optimization/}, 2025.
\newblock Version 9.12 per Google's citation page; experiments here use 9.15.

\bibitem[Perron et~al.(2023)Perron, Didier, and Gay]{perron2023cpsatlp}
Laurent Perron, Fr{\'e}d{\'e}ric Didier, and Steven Gay.
\newblock The {CP-SAT-LP} solver.
\newblock In \emph{29th International Conference on Principles and Practice of
  Constraint Programming (CP 2023)}, volume 280 of \emph{LIPIcs}, pages
  3:1--3:2. Schloss Dagstuhl -- Leibniz-Zentrum f{\"u}r Informatik, 2023.
\newblock \doi{10.4230/LIPIcs.CP.2023.3}.

\bibitem[Shen et~al.(2023)Shen, Song, Tan, Li, Lu, and
  Zhuang]{shen2023hugginggpt}
Yongliang Shen, Kaitao Song, Xu~Tan, Dongsheng Li, Weiming Lu, and Yueting
  Zhuang.
\newblock {HuggingGPT}: Solving {AI} tasks with {ChatGPT} and its friends in
  {Hugging Face}.
\newblock arXiv:2303.17580, 2023.

\bibitem[Topcuoglu et~al.(2002)Topcuoglu, Hariri, and Wu]{topcuoglu2002heft}
Haluk Topcuoglu, Salim Hariri, and Min-You Wu.
\newblock Performance-effective and low-complexity task scheduling for
  heterogeneous computing.
\newblock \emph{IEEE Transactions on Parallel and Distributed Systems},
  13\penalty0 (3):\penalty0 260--274, 2002.
\newblock \doi{10.1109/71.993206}.

\bibitem[Vieira et~al.(2003)Vieira, Herrmann, and Lin]{vieira2003rescheduling}
Guilherme~E. Vieira, Jeffrey~W. Herrmann, and Edward Lin.
\newblock Rescheduling manufacturing systems: A framework of strategies,
  policies, and methods.
\newblock \emph{Journal of Scheduling}, 6\penalty0 (1):\penalty0 39--62, 2003.
\newblock \doi{10.1023/A:1022235519958}.

\bibitem[Wang et~al.(2026)Wang, Liu, Feng, et~al.]{wang2026ontime}
Xinglin Wang, Zishen Liu, Shaoxiong Feng, et~al.
\newblock On time, within budget: Constraint-driven online resource allocation
  for agentic workflows.
\newblock arXiv:2605.06110, 2026.

\bibitem[Wu et~al.(2023)Wu, Bansal, Zhang, et~al.]{wu2023autogen}
Qingyun Wu, Gagan Bansal, Jieyu Zhang, et~al.
\newblock {AutoGen}: Enabling next-gen {LLM} applications via multi-agent
  conversation.
\newblock arXiv:2308.08155, 2023.

\bibitem[Yang et~al.(2025)Yang, Luo, Liu, Lou, and Chen]{yang2025bamas}
Liming Yang, Junyu Luo, Xuanzhe Liu, Yiling Lou, and Zhenpeng Chen.
\newblock {BAMAS}: Structuring budget-aware multi-agent systems.
\newblock arXiv:2511.21572, 2025.

\end{thebibliography}

\appendix
\section{Proof of the token-bucket proposition}
\label{app:proof}
\noindent\emph{Proof.} Suppose the first rejected request is sent at $\theta$, and let $\sigma\le\theta$ be the last time the bucket was full. The bucket never reaches its cap in $(\sigma,\theta]$, so its level before the request is $L+\frac{L}{60}(\theta-\sigma)$ minus the tokens sent earlier in $[\sigma,\theta]$, including requests processed before it at the same instant, and rejection means the tokens sent in $[\sigma,\theta]$ exceed $L\,(1+\frac{\theta-\sigma}{60})$. But $[\sigma,\theta]$ is covered by $\lfloor(\theta-\sigma)/60\rfloor+1\le 1+\frac{\theta-\sigma}{60}$ windows, each carrying at most $L$ tokens, a contradiction. $\square$

\section{Simulator, workflow and baseline details}
\label{app:sim}
\paragraph{Simulator.}
Realized input and output tokens are the estimates times independent log-normal factors ($\sigma=0.35$), and each request's duration, latency plus output tokens over speed, is multiplied by a log-normal jitter ($\sigma=0.15$).
A request whose realized tokens exceed the agent's context fails as an out-of-memory error (OOM) and is not billed.
A request is rejected with a 429, costing the caller 0.5\,s, if the provider's 60-second window (on realized tokens) or the agent's concurrency cannot admit it; a request larger than a whole minute's token limit fails unbilled.
The engine's shadow window records every dispatched request, including those answered with a 429.
An admitted attempt fails with probability $1-r_{at}$ and is billed.
A successful research task spawns two fact-check tasks with probability 0.3; each child depends on its parent, inherits its priority, takes the parent's output estimate as its input estimate, expects 200 output tokens and has no reliability floor, and the parent's existing children also wait for the new fact-checks.

\paragraph{Workflows.}
A seeded generator builds layered DAGs with 15 to 40 tasks in layers of 2 to 5 (the last layer may be smaller).
Each task gets one of six skills (research, code, math, writing, fact-check, summarize), with at least one research task per DAG; tasks after the first layer get one to three parents from earlier layers, log-uniform token estimates ($i_t\in[200,6000]$, $o_t\in[100,3000]$), a priority in 1--5 and a minimum reliability in $[0.6,0.9]$.

\paragraph{Online HEFT.}
Upward rank is the longest remaining path to the end of the DAG, using mean eligible durations.
HEFT decides at dispatch time and builds no static insertion schedule.
If its budget reserve would block every ready task while nothing is running, it runs the highest-ranked blocked task on its cheapest agent rather than stall.
Compared with our method it has no look-ahead, no global cost trade-off, no risk term and no use of priorities; it never waits for a faster agent to free up; and it enforces the budget through the reserve rather than as a constraint.

\section{Run history on the reporting seeds}
\label{app:history}
The reporting seeds produced six distinct sets of results while the system was being finished, and we list every run: seven executions in total (one re-execution after the bootstrap change reproduced identical per-run results), plus one diagnostic pass that read only solver statuses.
The first exposed infeasible re-solves (3 of 1{,}204), which we traced to pinning tasks in 429 back-off and fixed (Section~\ref{sec:rolling}).
After the second, we replaced a run-level bootstrap with the DAG-level one of Section~\ref{sec:setup}, on methodological grounds; this widened the intervals.
The next three runs followed changes not chosen from benchmark numbers: model speeds measured from real API calls, four bug fixes found while checking this paper's description against the code (the back-off cap, the 429 counter reset, HEFT reserving the cost of running tasks, which favors the baseline, and the fallback for tasks a failed re-solve leaves unplanned), and speeds re-measured with lower thinking settings after a live run showed hidden reasoning using up the output budget.
The last run used speeds measured with the per-model thinking settings of Section~\ref{sec:setup} and gives the numbers reported here.

For the follow-up of Section~\ref{sec:followup}, diagnostic runs on the tuning seeds, recorded in the pre-registration, found that the variants clearly faster than rolling also overspent more.
After the pre-registration, the $\lambda$ sweep ran on the tuning seeds, then the one pre-registered run on seeds 2000--2079, which gives the results reported here.
After release \code{\FuRef}, we re-executed that run once on unchanged code, only to time the solver (commit \code{\FuRerunCommit}); the results file came out byte-identical, and the rerun's solver times are all we use from it.
The commit that recorded the sweep also edited the new code twice: it rewrote the paid-reference bound in a form that computes the same value, and it changed how priced overspend reports the slack it used; neither changes any plan.
Seeds 0--19 were re-run with the two new options to check that the earlier results were unchanged; there priced overspend was \OldPricedFaster\,s faster than rolling, but those seeds produced the hypothesis and are not evidence for it.

\section{Live-run settings}
\label{app:live}
Each prompt includes the workflow goal and the parents' outputs, clipped to 16{,}000 characters in total and split evenly across parents.
Thinking is turned off where a model allows it and otherwise set to its lowest level, and \code{max\_tokens} is $\max(256,\lceil 3\,o_t\rceil)$, plus 1{,}024 for models whose thinking cannot be disabled (none ran here).

\end{document}